# Spatio-temporal forecasting of retaining wall deformation: mitigating error accumulation via multi-resolution ConvLSTM stacking ensemble

Jihoon Kim[a], Heejung Youn[*]

*Department of Civil and Environmental Engineering, Hongik University, 94, Wausan-ro, Mapo-gu, Seoul, 04066, Republic of Korea*

**Abstract.** This study proposes a multi-resolution Convolutional Long Short-Term Memory (ConvLSTM) ensemble framework that leverages diverse temporal input resolutions to mitigate error accumulation and improve long-horizon forecasting of retaining-structure behavior during staged excavation. An extensive database of lateral wall displacement responses was generated through PLAXIS2D simulations incorporating five-layered soil stratigraphy, two excavation depths (14 and 20 m), and stochastically varied geotechnical and structural parameters, yielding 2,000 time-series deflection profiles. Three ConvLSTM models trained at different input resolutions were integrated using a fully connected neural network meta-learner to construct the ensemble model. Validation using both numerical results and field measurements demonstrated that the ensemble approach consistently outperformed the standalone ConvLSTM models, particularly in long-term multi-step prediction, exhibiting reduced error propagation and improved generalization. These findings underscore the potential of multi-resolution ensemble strategies that jointly exploit diverse temporal input scales to enhance predictive stability and accuracy in AI-driven geotechnical forecasting.

**Keywords:** Convolutional long short-term memory, Retaining structure, Stacking ensemble, Time series prediction

## 1. Introduction

The deformation of retaining structures due to excavation induces settlement in adjacent areas and significantly affects structural stability, necessitating rigorous monitoring and management during construction [1, 2]. Although numerical [3] and back analysis [4] methods are commonly employed to predict and manage the deformation of these structures, accurately predicting deformation remains difficult due to inherent geotechnical uncertainties [5], and the process often requires considerable time and effort. As an alternative, many studies have used Artificial Intelligence (AI) to predict the behavior of retaining walls. Considering that the behavior of retaining structures evolves over time, specialized algorithms such as Recurrent Neural Networks (RNN), Long-Short Term Memory (LSTM), and Convolutional Long-Short Term Memory (ConvLSTM) have been employed [6, 7]. For instance, Tao et al. (2024) proposed a sequence-to-sequence ConvLSTM network to predict spatiotemporal wall deflections induced by deep excavations [8]. Liu et al. (2023) developed a prediction model that combines empirical mode decomposition (EMD) and RNN to physically predict the deformation of retaining structures in ultra-deep foundation pits [9], showing improved predictive accuracy over traditional empirical and numerical methods. Wang et al. (2024) introduced a deep learning framework based on spatiotemporal feature fusion to enhance excavation stability predictions, which was validated through extensive case studies and demonstrated superior performance and generalizability compared to existing models [10]. Zhao et al. (2021) proposed a Convolutional Neural Network (CNN)-based prediction method for concrete diaphragm wall deflections [11], achieving enhanced accuracy and efficiency relative to other RNN algorithms, and validating it through a deep excavation project in Suzhou, China. Samadi et al. (2024) compared diverse machine learning algorithms for predicting sidewall displacement in underground caverns and confirmed that the Backpropagation Neural Network (BPNN) provided the highest prediction accuracy [12]. Seo and Chung (2022) suggested a methodology to predict the lateral displacement of braced walls at each excavation stage, which employed a convolutional neural network in conjunction with an LSTM network, demonstrated its applicability for safety management and risk mitigation in construction [13].

[*] Corresponding author, Professor, Ph.D., E-mail: geotech@hongik.ac.kr

[a] Ph.D. Candidate, E-mail: jicaulayculkin@g.hongik.ac.kr

Although previous studies primarily focused on retaining structures, many researchers have explored the time-series behavior of various civil engineering systems and have demonstrated the superior performance of AI techniques. For instance, Mahmoodzadeh et al. (2022) developed and optimized an LSTM-based deep neural network model using 550 tunnel monitoring datasets to accurately predict tunnel wall convergence, outperforming traditional machine learning models such as Gaussian regression, support vector regression, K-nearest neighbor, and decision tree [14]. Yu et al. (2021) compared the performance of XGBoost, LSTM, and an Autoregressive Integrated Moving Average (ARIMA) model for time-series prediction in coastal bridge engineering, concluding that while all models could predict data trends over time, LSTM exhibited superior accuracy, particularly with improved data preprocessing and multi-step predictions [15]. Ma et al. (2021) introduced a novel deep learning approach employing a temporal graph convolutional network to predict slope deformation, taking into account spatial correlations between monitoring points, thereby providing a more comprehensive early warning system for slope failure compared to traditional models [16]. Mostafaei and Soleimani (2025) integrates Particle Swarm Optimization and neural network to develop a hybrid model that accurately predicts embankment dam displacement under dynamic loading using real-case data, demonstrating significantly improved performance over soft computing methods, making it a strong reference for neural network-based predictive modeling in geotechnical engineering [17]. Yang et al. (2019) presented a dynamic model utilizing time series analysis and LSTM network to predict landslide displacement in the Three Gorges Reservoir Area, demonstrating that the LSTM network outperformed static models by accurately capturing complex, nonlinear, and stepwise displacement patterns influenced by seasonal rainfall and reservoir water level fluctuations [18]. Li et al. (2024) proposed a decomposition, reconstruction, and optimization method using EMD and an optimized generalized RNN algorithm for predicting tunnel settlement deformation, demonstrating significantly enhanced accuracy and reliability over traditional models when applied to shallow tunnel data [19]. While these studies have successfully applied deep learning techniques to geotechnical engineering problems, most existing research focuses on single-step or short-term forecasts, which limits their effectiveness for long-term displacement forecasting.

In multi-step time series forecasting, two primary strategies are commonly employed: recursive (iterative) and direct method. The recursive method predicts one step ahead and recursively uses its own predictions as inputs for future steps [20]. While it is simple and flexible, this approach often suffers from error accumulation over extended prediction horizons. In contrast, the direct method predicts each fixed length of steps independently, either with separate models or a single multi-output model. This avoids the problem of error propagation but may struggle to capture temporal dependencies between prediction steps.

Su et al. (2025) enhanced multi-step prediction accuracy for dam displacement by incorporating a Chrono-initializer mechanism into an LSTM-based sequence-to-sequence model [21]. However, this approach does not capture spatial correlations, exhibits performance sensitivity to the length of the training sequence, lacks flexibility in output sequence length, and tends to respond insensitively to recent input data. Dietze et al. (2018) proposed an iterative near-term forecasting system for ecological applications [22]. While effective in short-term prediction, the approach incurs high computational costs due to frequent model updates and primarily focuses on short-range temporal patterns. Yang et al. (2024) introduced a deep learning framework leveraging an Adams-Bashforth-based time integration scheme to account for changes across multiple future time steps [23]. Nevertheless, their method tends to rely heavily on the distribution of the training data, limiting its ability to adapt to previously unseen patterns.

To overcome these limitations, this study suggests multi-resolution ConvLSTM ensemble framework designed to refine and enhance the multi-step accuracy in forecasting wall behavior during staged excavation. Fig. 1 illustrates the internal architecture of ConvLSTM network. Originally developed for predicting precipitation [24], the ConvLSTM network processes both spatial and temporal dynamics, making it applicable across domains such as video analysis [25], meteorological forecasting [26], and dynamic image processing [27].

ConvLSTM has the advantage of modeling spatio-temporal correlations within relatively uniform grids with low computational cost, without requiring integration with other networks or significant architectural modifications. In contrast, models such as vanilla LSTM, temporal convolutional network (TCN), and Transformers can capture temporal dependencies effectively, but they do not inherently encode spatial locality, making them less effective for modeling structured spatial patterns without additional design considerations. Spatio-temporal graph convolutional neural networks (ST-GCN) are well-suited for capturing spatio-temporal correlations in graph-structured (non-Euclidean) data; however, they are not specifically optimized for regular grid-based representations. Moreover, spatio-temporal Transformers have a limitation in that their computational cost increases quadratically with sequence length due to the attention mechanism. Therefore, ConvLSTM provides a more suitable and efficient framework for modeling spatio-temporal information in structured grid-based problems.

In this study, ConvLSTM models trained on varying temporal resolutions capture multi-scale temporal patterns, and their outputs are aggregated by a fully connected deep neural network serving as a meta-learner, resulting in a hierarchical ensemble architecture that improves the robustness and reliability of long-term forecasts. The effectiveness of the proposed approach is rigorously validated using a comprehensive dataset composed of synthetic data generated via PLAXIS2D simulations and field measurements collected from two real-world excavation projects.

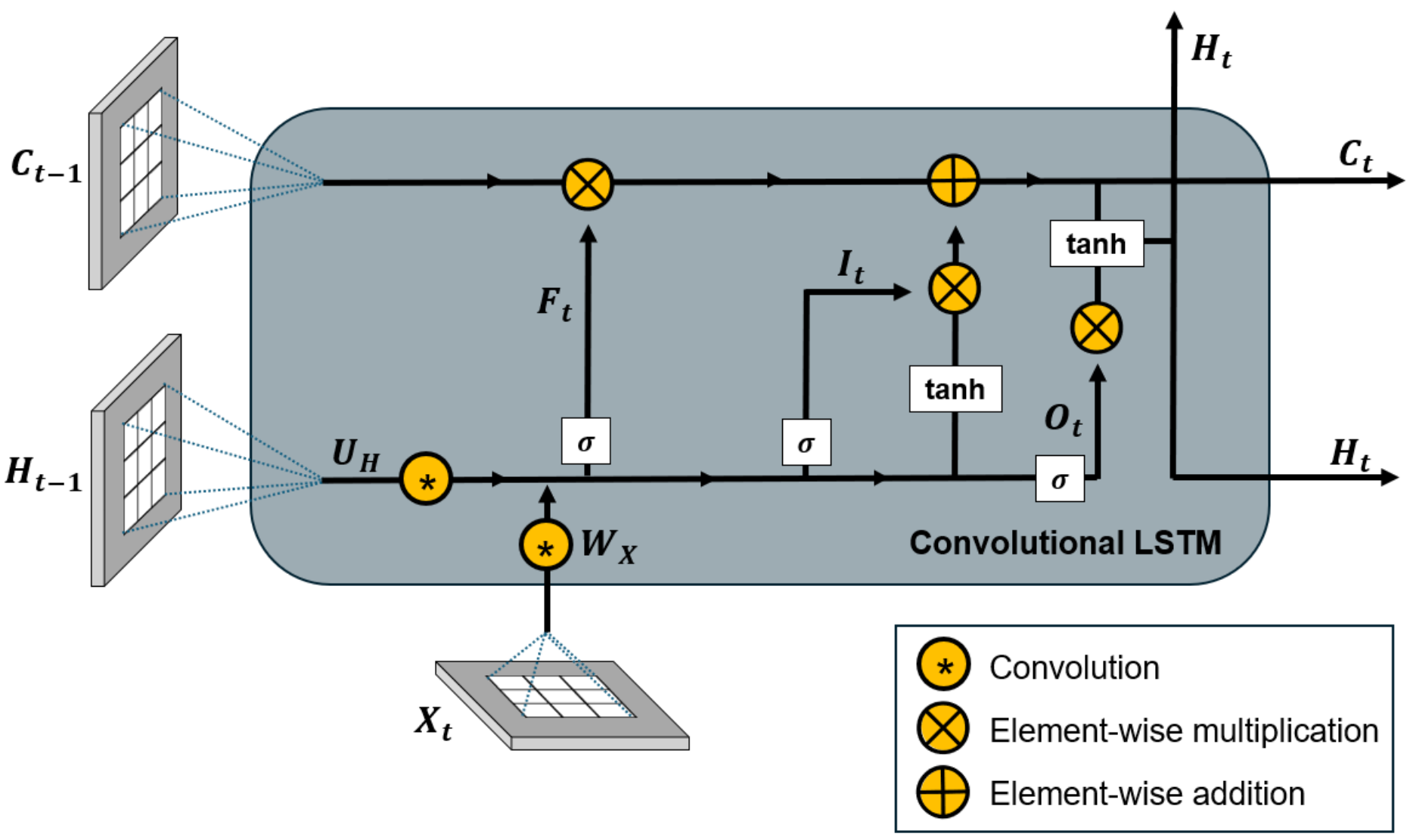

Figure 1. Conventional ConvLSTM network

## 2. Methodology

### 2.1 Proposed framework

#### 2.1.1 Overview

Fig. 2 presents an overview of multi-resolution ConvLSTM ensemble framework designed to refine and enhance the multi-step accuracy in forecasting wall behavior during staged excavation. The framework consists of three main stages: (1) data generation and preprocessing, (2) ConvLSTM-based multi-step prediction, and (3) stacking ensemble learning for prediction refinement.

A time-series deformation database is generated through numerical simulations using PLAXIS2D, capturing staged excavation-induced lateral displacements under varying soil conditions. The database comprises 2,000 excavation simulations, each represented as a sequential deformation profile. Multiple ConvLSTM models are trained using different input resolution (3, 6, and 10 resolutions) to capture short- and long-term deformation dependencies. To support this, a sliding window method is applied to reformat the simulation database into sequences (e.g., using a resolution of 3 yields 66,000 sequences). A recursive multi-step forecasting strategy is employed, where single-step predictions are iteratively used as inputs for future predictions. The outputs of different ConvLSTM models are aggregated using a stacking ensemble approach that applies deep learning (DL) based meta-learning technique. This framework enables the model to leverage diverse temporal dependencies captured by different input resolutions.

#### 2.1.2 Recursive forecasting

Multi-step forecasting is a critical aspect of time-series prediction [28], particularly in geotechnical applications where deformation patterns evolve progressively over time. To perform multi-step prediction, recursive forecasting, which is one of the fundamental strategies for long-term prediction, was employed [29].

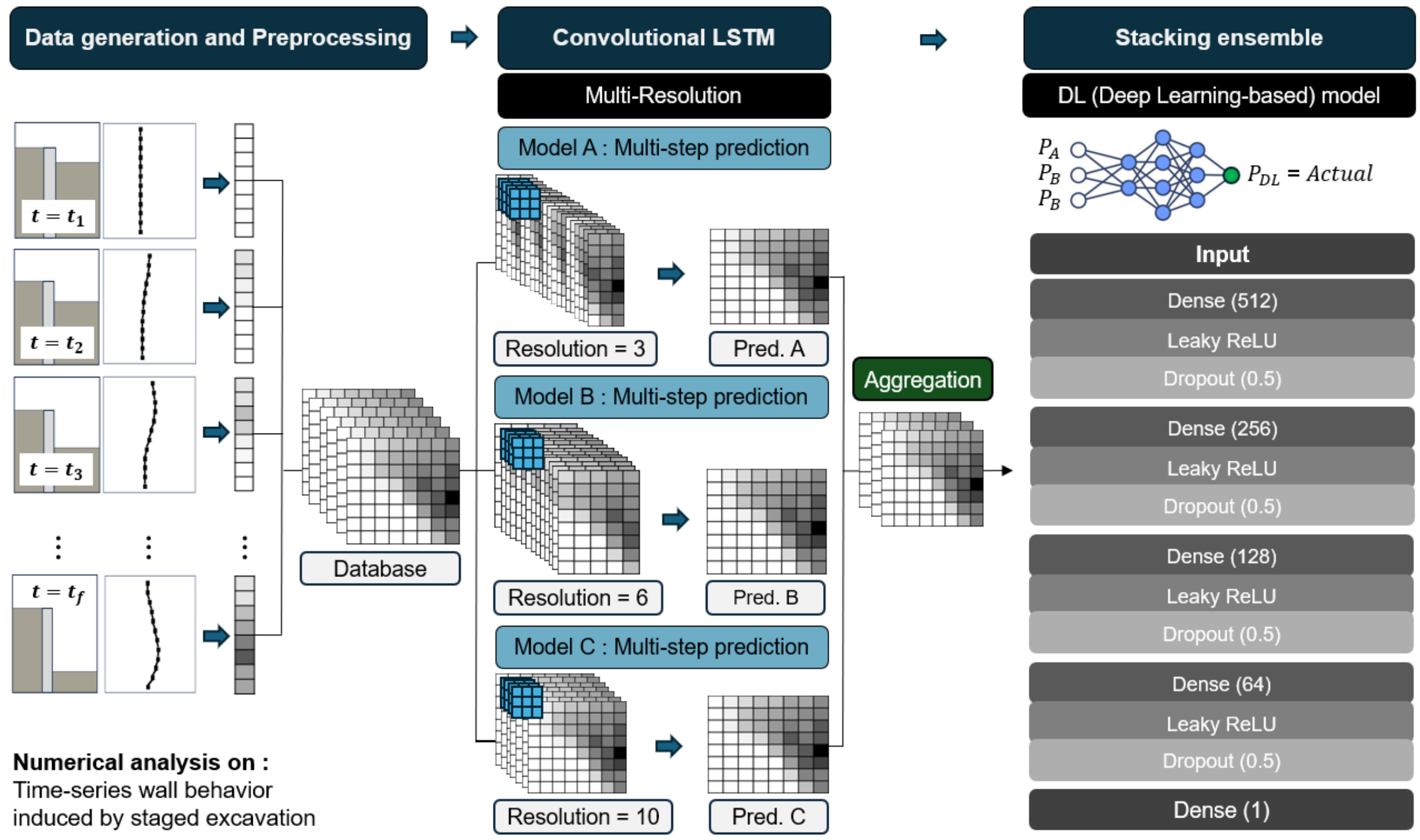

Figure 2. Overview of the proposed multi-resolution ConvLSTM ensemble framework

In recursive forecasting, the ConvLSTM model is initially trained for single-step forecasting, where it learns to predict next step based on a fixed-length sequence of past observations (i.e., resolution). Once trained, the model extends its forecasting capability by using its own predictions as inputs for future predictions. This process is repeated iteratively, allowing the model to generate multi-step forecasts beyond the original training window. Fig. 3 demonstrates this process in two parts. On the left, the training phase is shown where overlapping sequences are extracted from the time-series deformation data. Each sequence consists of a fixed number of past observations, determined by the resolution setting. On the right, the multi-step prediction process is illustrated where the model recursively generates predictions step-by-step, using previously predicted values as new inputs. It should be noted that each model is trained independently, and their outputs are combined with true labels to train the stacking ensemble model.

One of the key challenges in recursive multi-step forecasting is error propagation. As predictions are fed back into the model, small inaccuracies in early steps tend to amplify over time, degrading overall prediction reliability. This challenge was mitigated by employing a stacking ensemble approach [30]. In this study, multiple ConvLSTM models are trained with different resolutions. This will be discussed in the following section.

### 2.1.3 Stacking ensemble

The ConvLSTM-based network used in this study is designed for recursive single-step prediction with input resolutions of 3, 6, and 10. In a preliminary study, alternative multi-step forecasting strategies, including sequence-to-sequence and direct multi-output approaches, were also evaluated; however, the recursive forecasting strategy demonstrated superior predictive performance and was therefore adopted in this study. The model is implemented using a one-dimensional ConvLSTM (ConvLSTM1D) architecture to capture spatio-temporal dependencies along the wall depth.

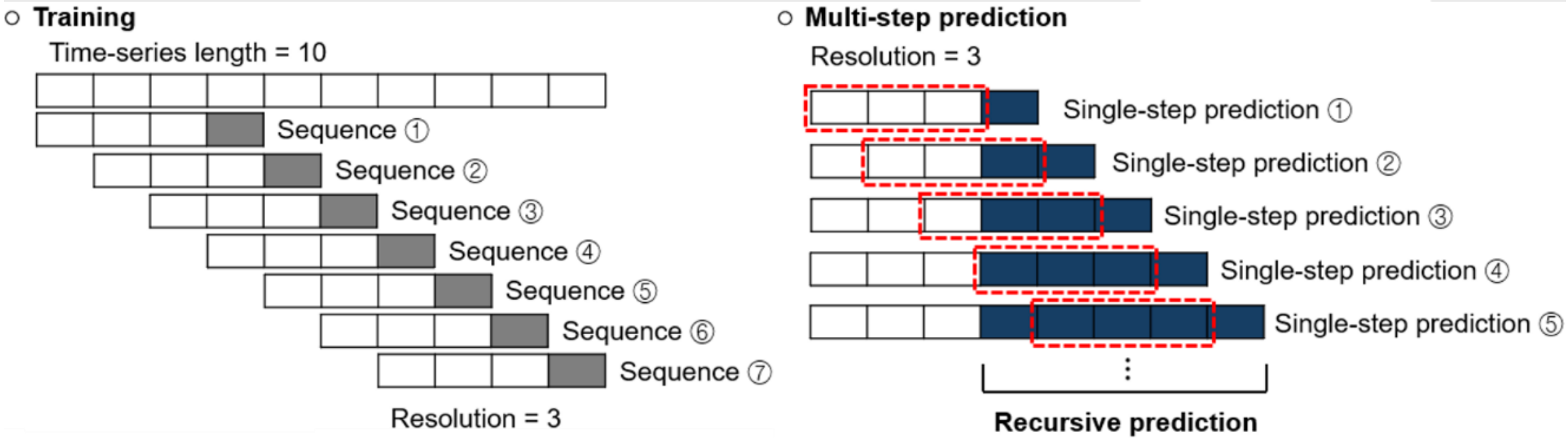

Figure 3. Example of recursive multi-step prediction process (resolution=3)

For each resolution, the model receives input sequences of shape (batch size, t, N, 1), where t denotes the temporal resolution (3, 6, or 10), N (=100) represents the number of spatial locations along the wall, and 1 corresponds to the feature dimension. The network consists of four stacked ConvLSTM1D layers with filter sizes of 128, 64, 32, and 8, respectively, each with a kernel size of 3, a stride of 1, and same padding applied to all layers. All ConvLSTM layers use the hyperbolic tangent (tanh) activation function. The first three ConvLSTM layers are configured with return sequences=True to preserve temporal information across layers, while the final ConvLSTM layer uses return sequences=False to aggregate the temporal features into a single representation. Dropout layers are applied after each of the first three ConvLSTM layers with rates of 0.2, 0.3, and 0.5, respectively, to mitigate overfitting.

The output of the final ConvLSTM layer is flattened into a one-dimensional vector and passed through a fully connected dense layer with N output units and a linear activation function, producing the predicted displacement profile across all spatial locations.

As the input propagates through the network, the output shape transitions from (t, 100, 128) in the initial ConvLSTM layer to (t, 100, 8) in the final layer. The final ConvLSTM output is flattened into a one-dimensional vector, where the output shape depends on the resolution: i.e., (None, 240) for t=3, (None, 480) for t=6, and (None, 800) for t=10. This representation is then processed through a fully connected dense layer with 100 units and a linear activation function, contributing 80,100 parameters to the model. The total number of trainable parameters varies with the resolution, with t=10 resulting is 467,332, the most complex configuration.

The multi-resolution ConvLSTM ensemble framework, as illustrated in Fig. 2, integrates multiple ConvLSTM models trained with different resolutions to improve multi-step prediction accuracy. This approach aims to construct an ensemble model, which is designed to capture the relationship between the multi-step predictions of base models and the ground truth using deep learning technique. Specifically, for the ensemble model, a fully connected deep neural network was trained using 10-step predictions from ConvLSTM models at each input sequence and the corresponding numerical simulation results. The model required training to be a larger number of parameters due to its multi-layer structure, increasing computational complexity but potentially capturing more intricate dependencies in the data. The network architecture consists of five sequential fully connected (dense) layers, comprising four hidden layers with 512, 256, 128, and 64 units, respectively, followed by a final output layer with a single unit. Although the input space of the meta-learner is low-dimensional, the network is designed to learn a nonlinear mapping that captures interactions among base-model predictions and corrects their systematic errors. Given the relatively deep architecture compared to the low-dimensional input space, there is a potential risk of overfitting. To address this, a dropout rate of 0.5 was applied after each dense layer to improve generalization and ensure robust learning. Leaky ReLU activation was applied to all layers, except for the output layer, to promote stable gradient propagation and mitigate the vanishing gradient problem during training. The model was optimized using the Adam optimizer with a learning rate of 0.001, and MSE was used as the loss function.

## 2.2 Data generation and preprocessing

### 2.2.1 Numerical simulations

Fig. 4 presents the numerical models developed in PLAXIS2D to simulate the lateral displacement behavior of a retaining wall subjected to staged excavation. The finite element models were designed to replicate excavation-induced deformation under realistic geotechnical conditions, incorporating a multi-layered soil profile (five layers), a retaining wall with struts, and staged excavation sequences. By stochastically varying geotechnical parameters and utilizing a range of wall and strut stiffness, a diverse set of time-series wall behavior data was obtained for model training. This will be discussed later.

The subsurface consists of five distinct layers, with a bedrock layer at the bottom underlying four soil layers. From the top, the thicknesses of these layers are 2, 4, 9, 9 and 16 m, respectively. Two excavation depths ($Z_f$) were considered: 14 m (Case A) and 20 m (Case B), with corresponding retaining wall length ($L_w$) of 18 and 26 m, respectively. It should be noted that the distinction between Case A and B is based on whether the wall tip is constrained. In Case B, the wall tip is embedded in the bedrock and thus constrained, whereas in Case A, it is not. The constraint condition of the wall tip significantly influences the time-series behavior of the wall during staged excavation, and by considering this condition differently, various behavioral responses can be obtained.

The retaining wall was modeled as a linear elastic material, and lateral support was provided by H-beam struts. The struts were installed at 3 m intervals along the depth of excavation, except for the first installation, which was placed 1 m below the ground surface in both cases. The excavation process was simulated in incremental phases, with soil layer being sequentially removed to replicate real-world excavation practices. The models assumed a plane strain condition, with sufficient lateral (80 m) and bottom (at least 14 m from wall end) boundary extent to prevent boundary effects from influencing wall deformations. The material properties of each layer and structural components are detailed in the following section.

### 2.2.2 Staged excavation

The staged excavation process was simulated using 0.5 m incremental excavation steps, replicating real-world construction practices and the sequential activation of lateral support system (strut). This methodology ensured that excavation-induced stress redistribution and wall deformations were realistically captured. The detailed excavation sequences for Case A (final depth of 14 m) and Case B (final depth of 20 m) are presented in Table 1.

As previously described, the first strut was installed at a depth of 1.0 m, with subsequent struts placed at 3.0 m intervals to maintain structural stability during excavation. The analysis assumed that no additional jacking forces were applied during strut installation, reflecting conventional field conditions. In both cases, the retaining wall was installed in the first phase before any excavation commenced. Excavation for Case A proceeded through 29 sequential phases until reaching a final depth of 14.0 m, while Case B followed 41 phases to a final depth of 20.0 m, with the same staged strut installation.

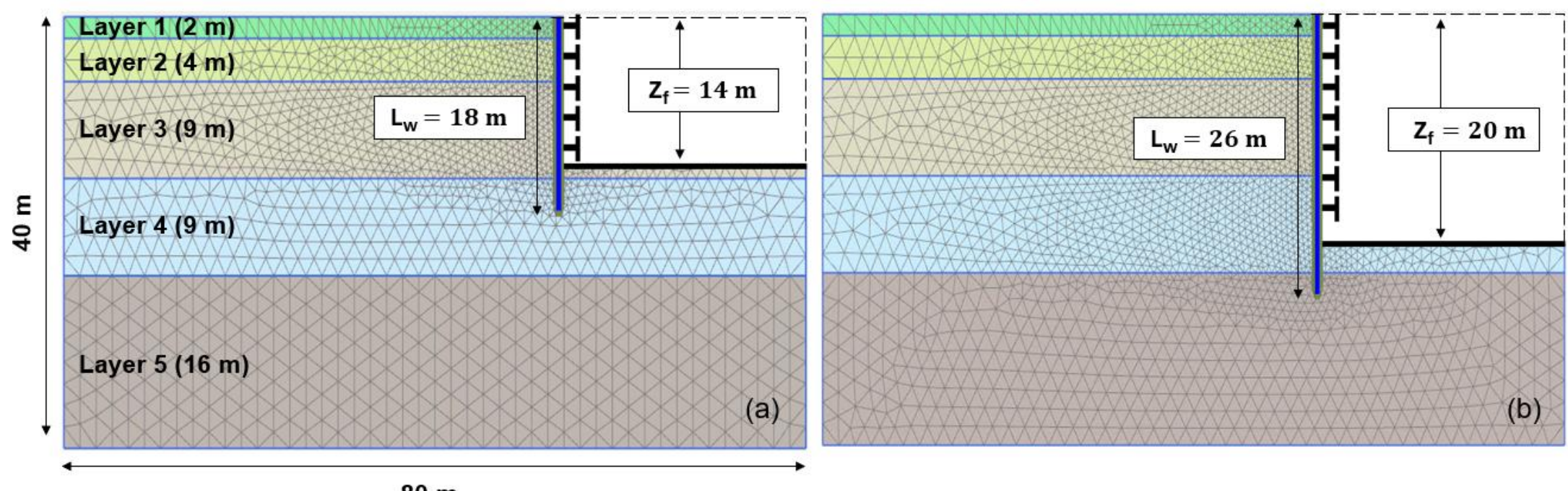

Figure 4. Numerical models of retaining walls for training with excavation depths of: (a) 14 m (Case A) and 20 m (Case B)

Table 1. Excavation depth of strut installation for two cases

| Phase No. | Case | | Excavation depth (m) | Installation |
|---|---|---|---|---|
| 1 | A | B | 0.0 | Retaining wall |
| 4 | A | B | 1.5 | 1st strut at 1.0 m |
| 10 | A | B | 4.5 | 2nd strut at 4.0 m |
| 16 | A | B | 7.5 | 3rd strut at 7.0 m |
| 22 | A | B | 10.5 | 4th strut at 10.0 m |
| 28 | A | B | 13.5 | 5th strut at 13.0 m |
| 29 | A | | 14.0<br>(final depth for Case A) | N/A |
| 34 | | B | 16.5 | 6th strut at 16.0 m |
| 40 | | B | 19.5 | 7th strut at 19.0 m |
| 41 | | B | 20.0<br>(final depth for Case B) | N/A |

### 2.2.3 Input material parameters

The geotechnical properties of the subsurface materials were assigned based on five soil layers, consisting of four soil layers using the Hardening Soil model and one bedrock layer modeled by the Mohr-Coulomb model. The Hardening soil model was chosen for its ability to capture stress-dependent stiffness, strain-hardening behavior, and stress path effects [31], while Mohr-Coulomb model was applied to the bedrock, assuming a linear elastic-perfectly plastic response appropriate for stiff materials under relatively low normal stress conditions [32].

To ensure the generalizability and robustness of the predictive framework, a stochastic parameter assignment approach was employed, incorporating normally distributed random sampling for key geotechnical properties. Specifically, the total unit weight ($\gamma_t$), internal friction angle ($\phi'$), cohesion ($c'$), and elastic modulus ($E_{ref}$) were assigned as random variables following a normal distribution. The mean and standard deviation of each parameter were predefined based on empirical data [33], as summarized in Table 2.

Table 2. Soil input parameters and constitutive models for each layer

| Layer | Constitutive model | Input parameters | Mean value | Standard deviation |
|---|---|---|---|---|
| 1 | Hardening Soil | $\gamma_t$ (kN/m$^3$) | 18.0 | 2.0 |
| | | $c'$ (kPa) | 1.8 | 0.5 |
| | | $\phi'$ (°) | 28.0 | 2.5 |
| | | $E_{50}^{ref}$ (kPa) | 15,000 | 3,000 |
| | | $e_{init}$ | 0.6 | N/A |
| 2 | Hardening Soil | $\gamma_t$ (kN/m$^3$) | 19.0 | 2.0 |
| | | $c'$ (kPa) | 5.0 | 1.5 |
| | | $\phi'$ (°) | 28.0 | 3.0 |
| | | $E_{50}^{ref}$ (kPa) | 30,000 | 6,000 |
| | | $e_{init}$ | 0.5 | N/A |
| 3 | Hardening Soil | $\gamma_t$ (kN/m$^3$) | 20.0 | 2.0 |
| | | $c'$ (kPa) | 20.0 | 5.0 |
| | | $\phi'$ (°) | 32.0 | 3.0 |
| | | $E_{50}^{ref}$ (kPa) | 40,000 | 8,000 |
| | | $e_{init}$ | 0.4 | N/A |
| 4 | Hardening Soil | $\gamma_t$ (kN/m$^3$) | 21.0 | 2.0 |
| | | $c'$ (kPa) | 50.0 | 10.0 |
| | | $\phi'$ (°) | 32.0 | 3.0 |
| | | $E_{50}^{ref}$ (kPa) | 60,000 | 15,000 |
| | | $e_{init}$ | 0.4 | N/A |
| 5 | Mohr-Coulomb | $\gamma_t$ (kN/m$^3$) | 23.0 | 2.0 |
| | | $c'$ (kPa) | 80.0 | 20.0 |
| | | $\phi'$ (°) | 35.0 | 3.0 |
| | | $E'_{ref}$ (kPa) | 1,000 | 200 |
| | | $e_{init}$ | 0.01 | N/A |

This approach introduces natural variability into the numerical database, allowing the predictive model to learn a broader range of soil behavior patterns, which is critical for improving generalization across different site conditions. Examples of normally distributed properties for Layer 1 are shown in Fig. 5. The additional elastic modulus components, $E_{oed}^{ref}$ and $E_{ur}^{ref}$, in the Hardening Soil model were assigned based on predefined correlations: $E_{oed}^{ref} = E_{50}^{ref}$ and $E_{ur}^{ref} = 3 \cdot E_{50}^{ref}$, respectively. Fixed initial void ratios were used for all layers.

The structural components in the numerical model, including the retaining wall and struts, were assigned fixed material properties, unlike the soil layers where normally distributed parameters were used. However, similar to the randomized selection of soil properties, the input parameters for the retaining wall and struts were randomly selected from a predefined dataset. This ensured that different wall thickness and strut cross-sectional areas were considered in the simulations, allowing the model to learn from diverse structural conditions. The input parameters for the retaining wall and struts are summarized in Table 3 and 4.

For the retaining wall, 6 representative flexural stiffness (EI) values were defined. Wall thickness was the critical variable as EI directly dependent on thickness, influencing the wall deformation response. The flexural stiffness used in this study ranges from 100 to 1,200 MN·m$^2$/m with corresponding wall thickness of 0.35 and 1.20 m. For the strut, a total of 6 representative axial stiffness values were selected based on the specifications of H-beams commonly used in Korea [34]. The axial stiffness values ranged from 670 to 2,194 MN, reflecting the structural diversity in support system.

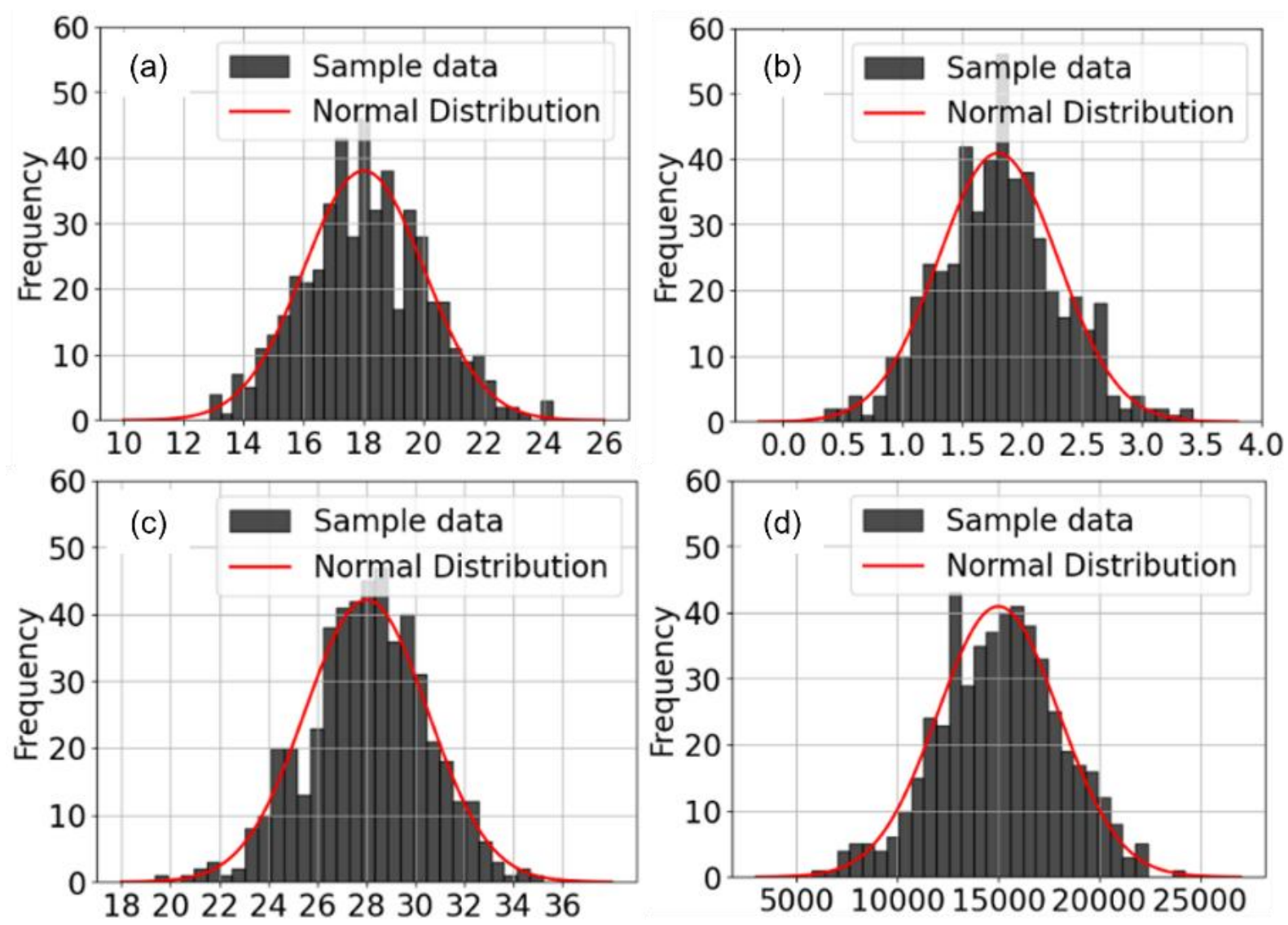

Figure 5. Example of statistical distribution of soil input parameters (Layer 1), (a) total unit weight, (b) cohesion, (c) internal friction angle, and (d) elastic modulus

Table 3. Randomly selected input parameters for the retaining wall

| Type | A | B | C | D | E | F |
|---|---|---|---|---|---|---|
| Flexural stiffness (MN·m$^2$/m) | 100 | 200 | 400 | 600 | 800 | 1,200 |
| Thickness (m) | 0.35 | 0.49 | 0.70 | 0.85 | 0.98 | 1.20 |

Table 4. Randomly selected input parameters for struts

| Type | A | B | C | D | E | F |
|---|---|---|---|---|---|---|
| Cross-sectional area (mm$^2$) | 3,270 | 4,680 | 5,620 | 7,240 | 8,820 | 10,700 |
| Axial stiffness (MN) | 670 | 959 | 1,152 | 1,484 | 1,808 | 2,194 |
| Elastic modulus (MPa) | 200,500 | | | | | |

## 2.3 Database construction

To systematically generate a large-scale training database, the entire numerical simulation process, including model execution, stepwise data extraction, and preprocessing, was fully automated using the PLAXIS2D API and Python scripting. Lateral wall displacements were extracted from designated monitoring points spaced at 0.5 m intervals along the retaining wall, covering its entire height.

In Case A, 37 monitoring points recorded displacement data across 29 excavation phases, while in Case B, 53 monitoring points captured deformation over 41 phases. A total of 1,000 numerical simulations were conducted for each case, yielding a database of 2,000 simulations. To mitigate issues such as vanishing and exploding gradients during model training, displacement values were systematically stored in meters to maintain a numerically stable range. This unit selection is particularly advantageous given that excavation-induced wall displacements in the elastic range rarely surpass 1.0 m. By adopting meters as the standard unit, numerical conditioning is improved, facilitating more stable gradient propagation and ultimately promoting stable model convergence and reliability.

To ensure that the predictive model could handle walls of different lengths while maintaining a consistent input format, the extracted displacement profiles were normalized to a fixed spatial resolution. The raw displacement values from numerical simulations were spline-interpolated to 100 evenly spaced points along the wall length, standardizing the database and allowing for a uniform input structure regardless of excavation depth. This preprocessing step ensured that all simulations contributed the same number of spatial features to the training database, preventing discrepancies due to the wall length variations.

Following spatial normalization, the interpolated displacement profiles were structured for training. The collected data were stacked along the batch axis, effectively grouping them for unified processing. Finally, the entire database was reshaped and restructured according to the predefined resolutions. Specifically, for Case A, the resulting data had dimensions of (1,000, 29, 100, 1), and for Case B, dimensions of (1,000, 41, 100, 1), where the first dimension represents the batch size, the second represents the time-series length (excavation phases), the third represents the interpolated wall length (100 points), and the fourth represents the number of features (displacement in this study). The database was then reformatted into (N, 3, 100, 1), (N, 6, 100, 1) and (N, 10, 100, 1) for different temporal resolutions based on sliding window method, where N denotes the number of sequence samples. For example, with a resolution of 3, a total of 66,000 sequences were generated. Then, the reformatted dataset was divided into train, validation, and test sets at a ratio of 7:2:1, resulting in 46,200 training sequences, 13,200 validation sequences, and 6,600 test sequences, respectively.

It should be noted that each resolution corresponds to a 0.5 m change in excavation depth, thus, resolutions of 3, 6, and 10 indicate that wall deformation over excavation depths of 1.5 m, 3.0 m, and 5.0 m, were respectively used to generate predictions. Fig. 6 illustrates the time-series lateral displacement profiles of the retaining wall obtained from the numerical simulations for Case A and Case B. As excavation progresses, lateral displacements increase, with Case B exhibiting ordinarily greater overall displacement than Case A, reflecting the effect of increased excavation depth.

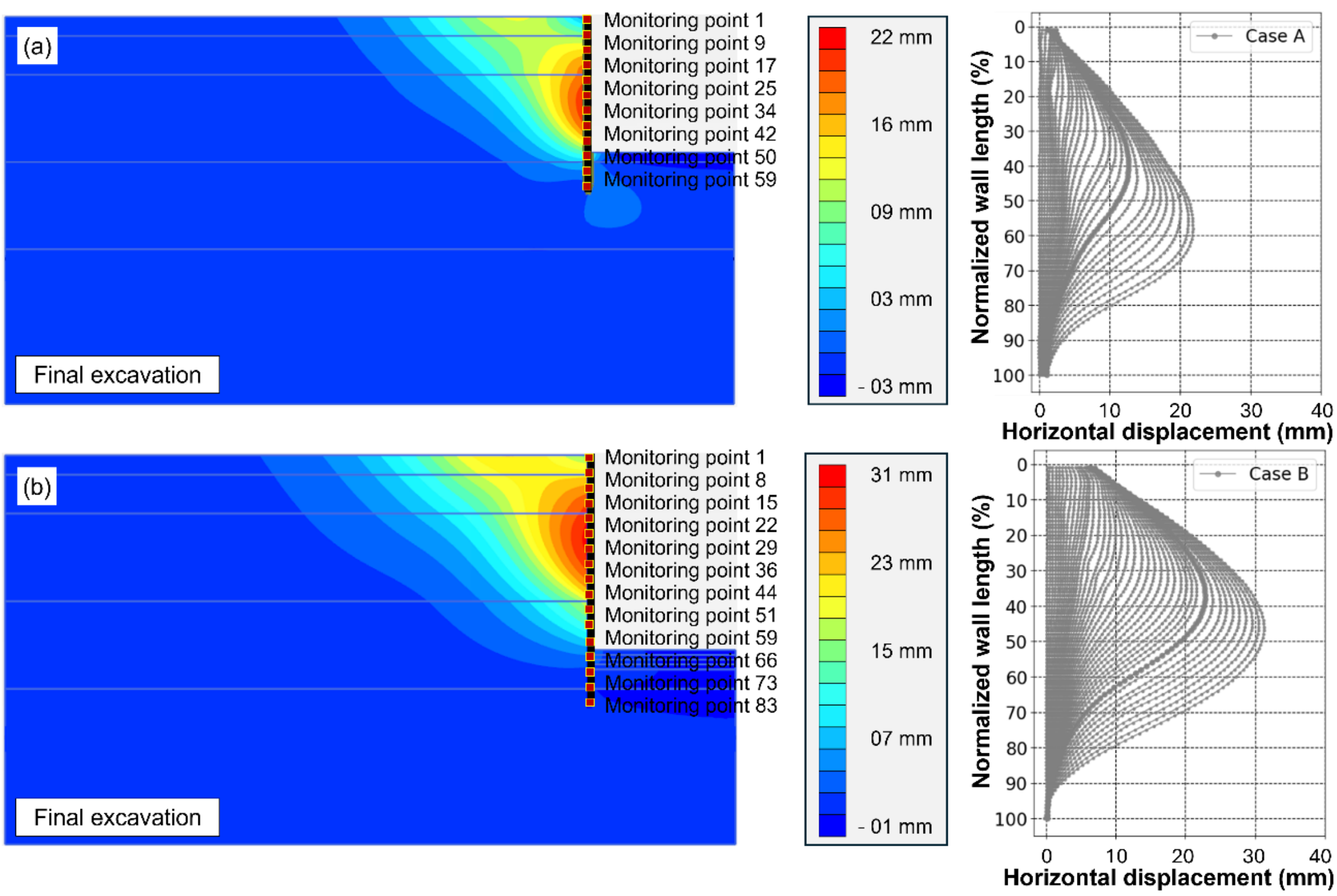

Figure 6. Numerical simulation results and the obtained wall displacement of (a) Case A and (b) Case B

## 3. Results and Discussion

### 3.1 Multi-step prediction for numerical analysis

The overall multi-step prediction performance of standalone ConvLSTM models and stacking ensemble model (DL model) is presented in Fig. 7, which illustrates the trends in Mean Absolute Error (MAE), Coefficient of Determination ($R^2$), and Index of Agreement (IoA) [35] across different prediction steps. These results were derived by averaging the MAE, $R^2$, and IoA values computed for each sequence within the test set (10% of total database). The results highlight the error accumulation inherent in recursive forecasting and demonstrate the effectiveness of the stacking ensemble approach in mitigating this issue.

The standalone ConvLSTM models (A, B, and C) exhibit a clear increase in MAE as the prediction horizon extends. The error accumulation is most pronounced in Model C (resolution=10), followed by Model A (resolution=3) and Model B (resolution=6). The $R^2$ values (Fig. 7(b)) shows significant drop beyond 3$^{rd}$ prediction step for all models, with the Model C experiencing the steepest decline, and the values fall below 0 at 7$^{th}$ prediction step for all models, confirming that long-term predictions become increasingly unreliable due to cumulative errors. These results emphasize the limitations of using standalone models for extended multi-step prediction.

In contrast, the DL model demonstrates significant improvements over the standalone models. The DL model consistently yields lower MAE while maintaining higher $R^2$ and IoA values across all prediction steps. It maintains an IoA above 0.94 even at 10$^{th}$ prediction step, indicating its ability to stabilize long-term predictions and suppress error propagation. This observation confirms that the stacking ensemble approach significantly enhances the predictive accuracy of multi-step forecasting.

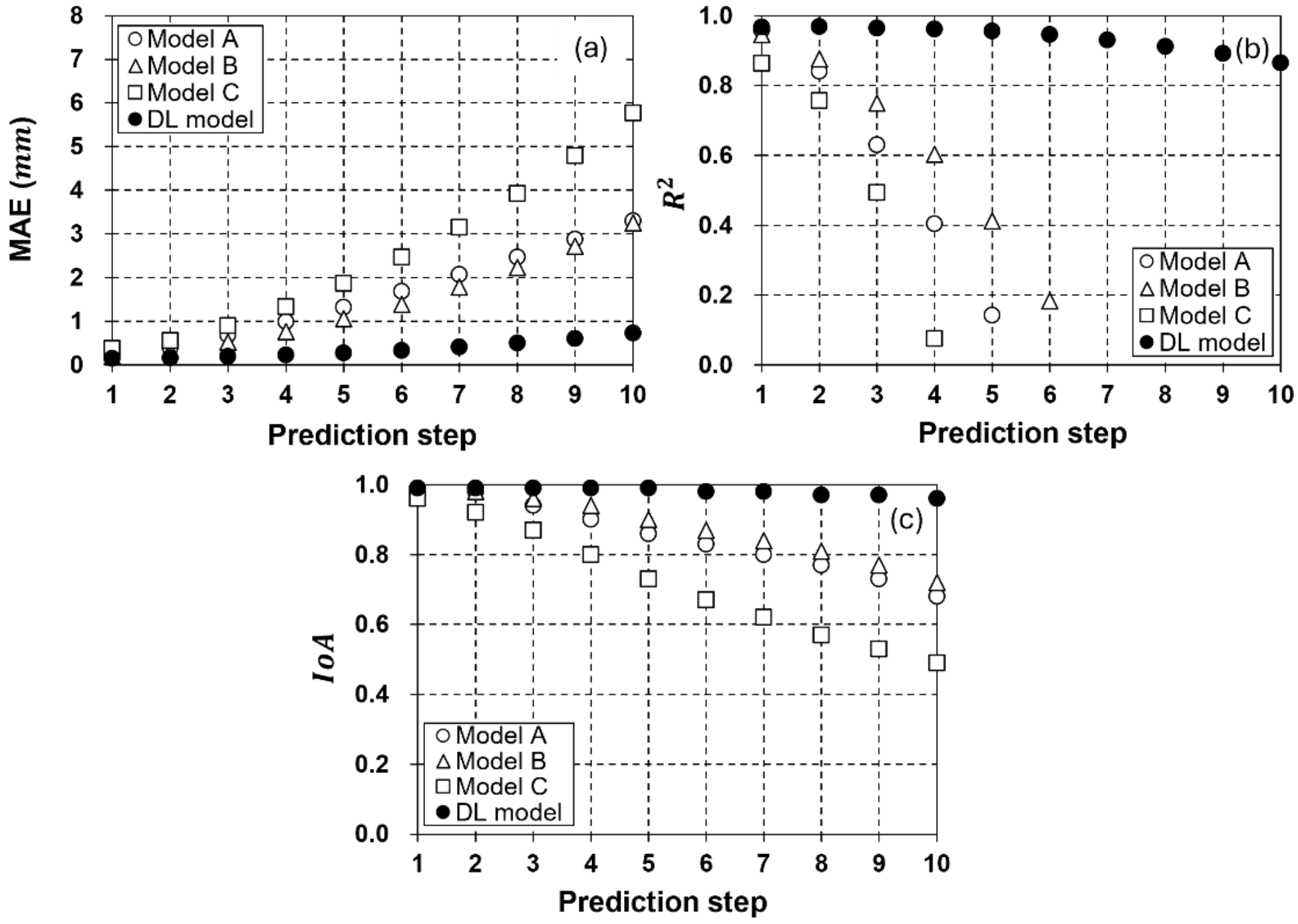

Figure 7. Averaged multi-step prediction performance of standalone ConvLSTM and stacking ensemble models for numerical simulation results: (a) MAE, (b) $R^2$, and (c) IoA

Fig. 8 presents an example of multi-step prediction using standalone models and ensemble model for forecasting lateral wall displacements during staged excavation. For numerical simulation results, Model B is identified as the most accurate model among standalone models. Accordingly, stepwise IoA values of Model B and DL model are presented and compared, highlighting the effectiveness of the proposed framework in improving multi-step predictive accuracy.

The leftmost plot represents the numerical simulation results from excavation depth of 0-4.5 m, which were used as input sequence to predict the subsequent seven steps. The standalone models (Model A, B, and C) employed different resolution configurations, utilizing the most recent 3, 6, and 10 steps, respectively, to generate predictions. The subsequent plots compare actual displacement profile (numerical simulation results) with predictions from four different models. As excavation progresses, the lateral displacement increases, reflecting the excavation-induced wall deformation. While all models initially exhibit high accuracy (IoA > 0.97), deviations from actual displacement become more pronounced for the standalone models as the prediction horizon extends, consistent with earlier findings. For short-term predictions (prediction step of 1-3), all models closely follow the actual displacement trend. However, error accumulations intensify in the models over longer horizons, leading to overestimation by Models A and B and underestimation by Model C. In contrast, DL model effectively integrates the predictive attributes of the standalone models, yielding more accurate multi-step forecasting. Notably, the DL model achieved the highest accuracy, maintaining IoA of 0.98 at 10th prediction step, demonstrating its reliability in predicting excavation-induced wall deformations under FEM analysis. In contrast, Model B exhibited a significantly lower IoA of 0.62 at the same step.

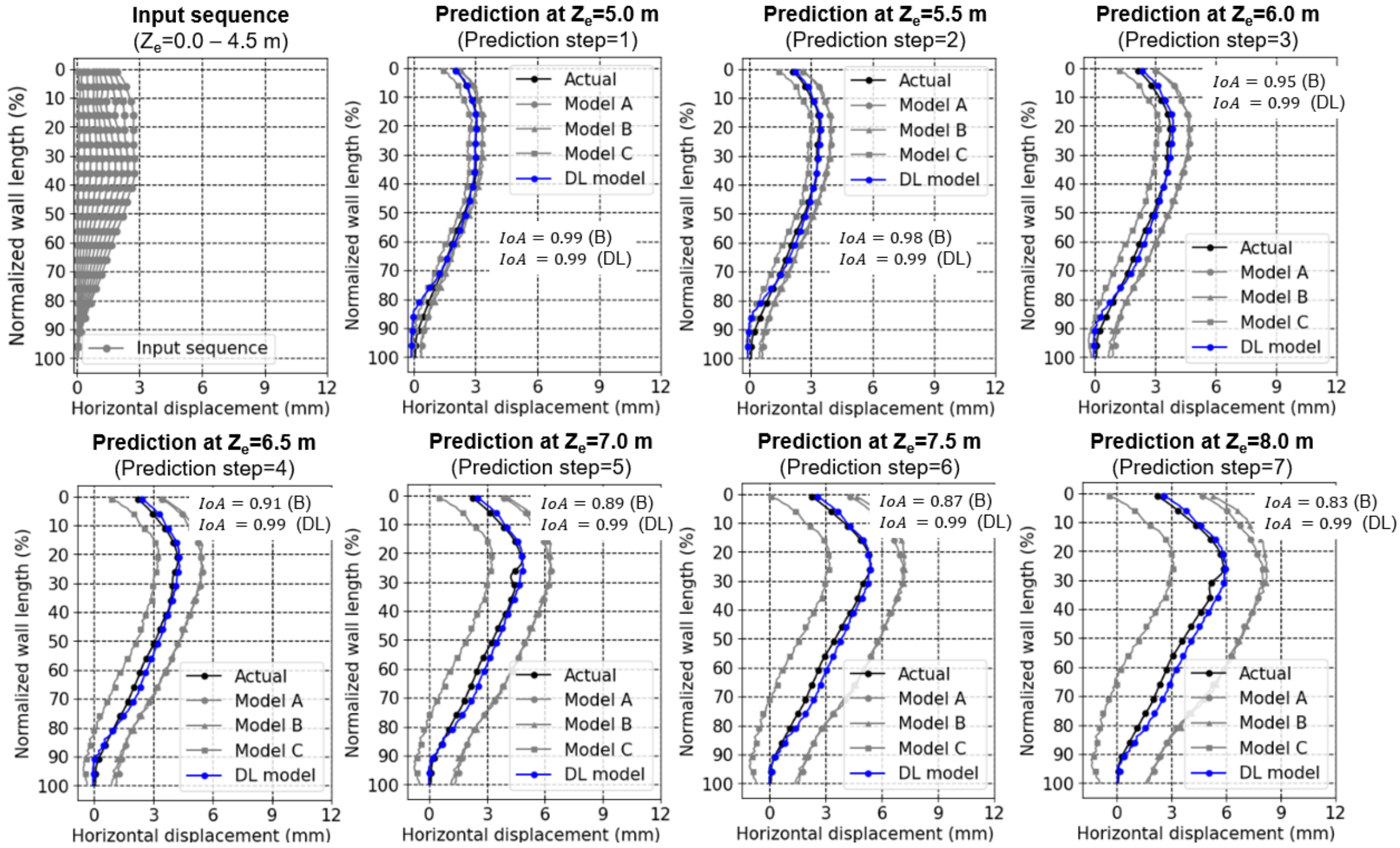

Figure 8. Multi-step predictions of standalone ConvLSTM and ensemble models using input sequence from excavation depth of 0-4.5 m

Fig. 9 illustrates another example of multi-step prediction, following the same forecasting process as Fig. 8 but employs a shifted input sequence (excavation depth of 4.5-9.0 m), enabling for a comparative assessment under different prediction conditions. The overall trends in model performance remain consistent with Fig. 8. The standalone models again exhibit increasing deviation over longer prediction horizons, with Model C demonstrating the highest error accumulation. Models A and B continue to overestimate displacement, while Model C tends to underestimate it in later steps. The DL model once again outperformed the standalone models, achieving the most accurate multi-step predictions, maintaining IoA of 0.98 at 10th prediction step. A similar trend is observed across other input sequences, confirming that stacking ensemble framework significantly enhances long-term prediction accuracy.

To better explain the internal decision-making process of the ensemble model, we investigated the contribution of each standalone model (Model A, B, and C) to the ensembled prediction at each forecasting step. The Shapley Additive exPlanations (SHAP) [36] summary plots presented in Fig. 10, visualize how the contributions of the standalone models to the final ensemble prediction evolve over the course of 10 prediction steps, based on the entire test set (approximately 6,600 sequences). Each subplot corresponds to a specific prediction step and illustrates the SHAP values associated with each standalone model's prediction, where the horizontal axis represents the contribution to the ensemble output and the color encodes the magnitude of the standalone model's prediction (i.e., feature value). These plots allow for a detailed inspection of not only how much influence each standalone model has on the ensemble decision, but also how that influence varies across the prediction horizon.

In these plots, larger absolute SHAP values indicate greater impact on the ensemble output, while the sign of the SHAP value reflects whether the base model's prediction increased or decreased the final forecast. Across all prediction steps, several consistent patterns are observed. Most notably, Model C exhibits the largest spread in SHAP values throughout all prediction steps, indicating that its predictions contribute most significantly to the ensemble model's output. Model A generally shows intermediate levels of influence, while Model B consistently displays the smallest SHAP value distribution, suggesting it plays a more limited role in the final prediction. This ordering of influence (Model C > A > B) remains stable across the forecast horizon. In addition, as the prediction step increases, the overall SHAP distributions tend to widen for all models. This broadening can be interpreted as an increase in predictive uncertainty; that is, the ensemble model becomes more sensitive to the input predictions as the temporal distance from the conditioning input grows. Interestingly, although Model B achieves the highest prediction accuracy among standalone models (refer to Figs. 7 and 8), its influence on the ensemble prediction is consistently the lowest. Conversely, despite being the weakest individual performer, Model C contributes most heavily within the ensemble framework. This apparent discrepancy might be explained by the training mechanism of the ensemble model, which is optimized not for the accuracy of individual inputs, but rather for the overall reduction of ensemble prediction error. In this context, the ensemble may find Model C useful not for its correctness, but for its ability to provide diverse or corrective signals that compensate for the systematic biases of other models.

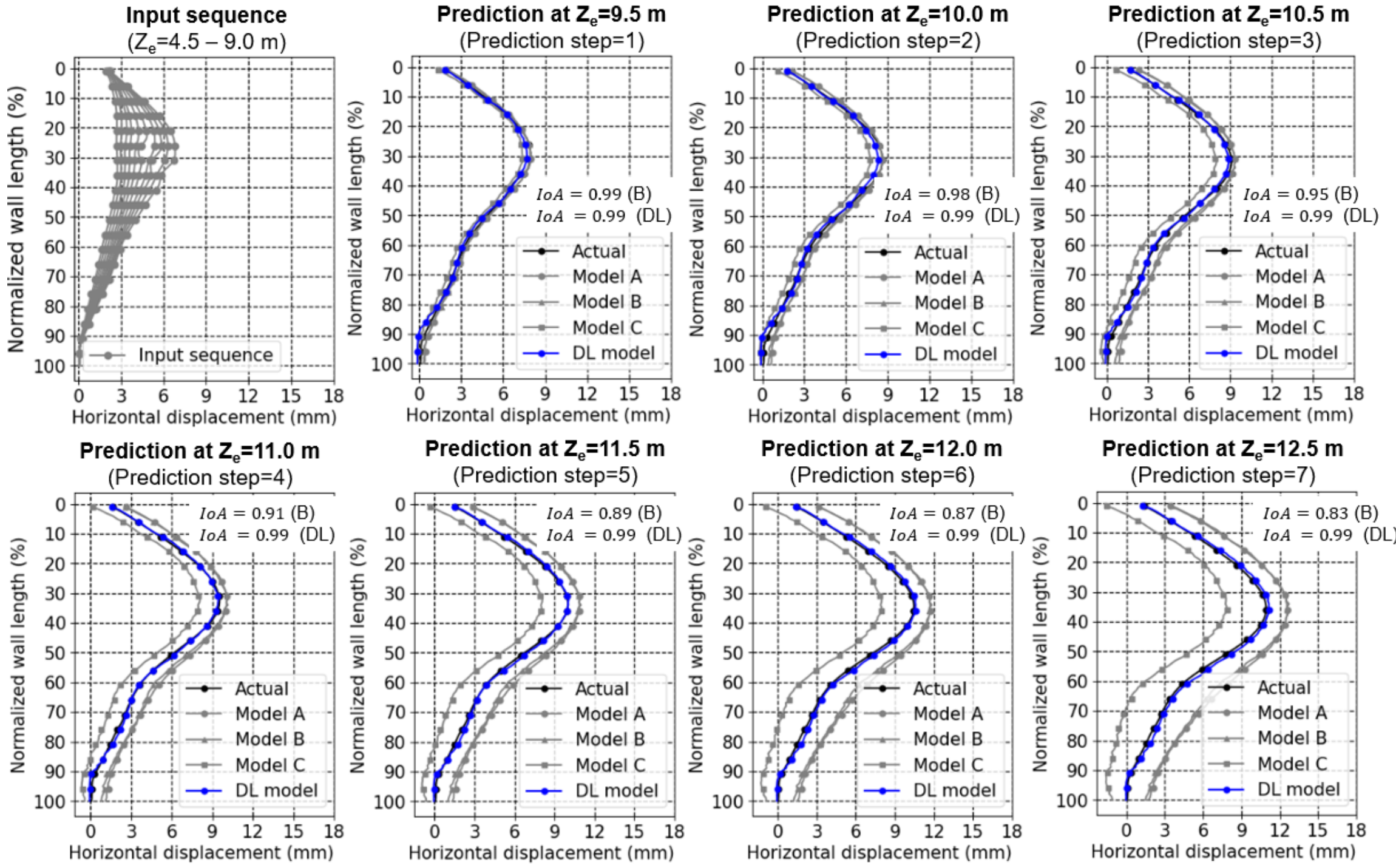

Figure 9. Multi-step predictions of standalone ConvLSTM and ensemble models using input sequence from excavation depth of 4.5-9.0 m

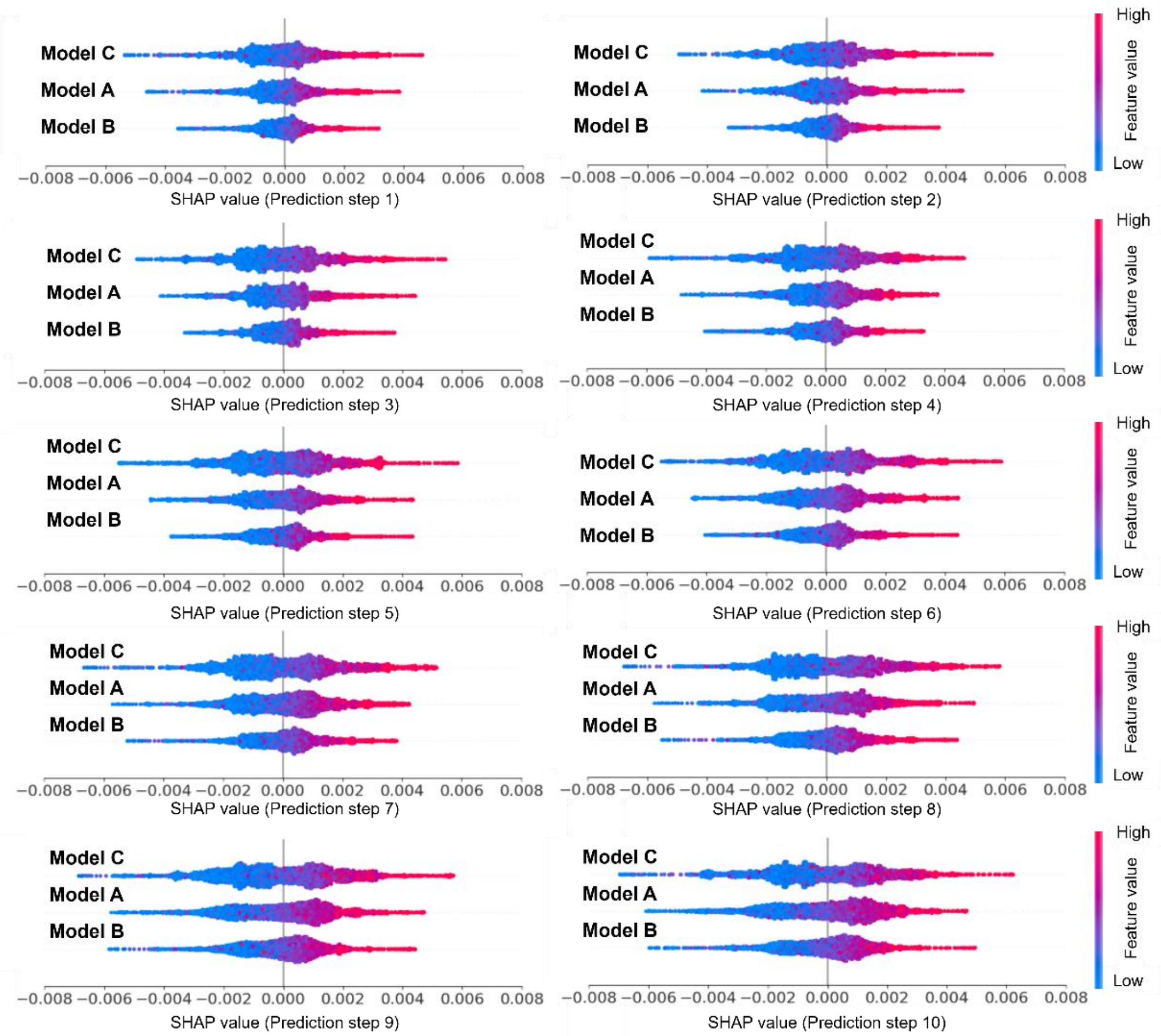

Figure 10. SHAP value distributions for Model A, B, and C across prediction steps 1 to 10, showing their contributions to the ensemble model's predictions

## 3.2 Validation with field measurements

To evaluate the practical applicability of the developed framework, the multi-step predictive accuracy of standalone and ensemble models was assessed using field measurements from two urban inland excavation sites in South Korea. Fig. 11 shows the lateral displacement profiles at discrete depths for Sites A and B during staged excavation.

At Site A, five measurements were recorded at excavation depths ($Z_e$) of 0, 3, 5, 8, and 10 m, with a maximum displacement of 11 mm. The wall length is 12 m, and displacement profiles were interpolated and resampled along the normalized wall length to ensure consistent spatial resolution. To satisfy the input requirements of the model, the measurements at discrete excavation depths were further interpolated at 0.5 m intervals. The profile (Fig. 11(a)) shows pronounced displacement increases, particularly in the upper half.

At Site B, seven measurements were collected at $Z_e$ of 0, 2.5, 5, 7.5, 11, 14.5, and 17 m, with a maximum displacement of 17 mm. The wall length is 20 m, and the same spatial normalization and temporal interpolation procedures were applied. Unlike Site A, Site B exhibits pronounced displacement in the central portion, while it is restrained at the top by structural supports and at the bottom by tip restraint from the underlying soft rock layer.

Based on these measurements, predictions are made using the deformation history up to the $N^{th}$ excavation stage to estimate wall displacement at the subsequent stage $(N+1)^{th}$. Because the framework requires a deformation history spanning 5.0 m (10 steps), the available target excavation depths for prediction at Site A are 8.0 and 10.0 m. Specifically, the wall behavior at 8.0 m is predicted using measurements up to 5.0 m, corresponding to a forward prediction over an excavation advance of 3.0 m (prediction step=6). Similarly, the deformation at 10.0 m is predicted based on measurements up to 8.0 m, also representing a 2.0 m excavation advance (prediction step=4). Using the same approach, the available target excavation depths for evaluation at Site B are 7.5, 11.0, 14.5, and 17.0 m. Specifically, the wall behavior at each target depth is predicted using the preceding deformation history, corresponding to excavation advances of 2.5, 3.5, 3.5, and 2.5 m (prediction steps=5, 7, 7, and 5), respectively.

Overall, the numerical simulation data used for training shows a uniformly increasing displacement trend, whereas the field data in Fig. 11 exhibit irregular patterns due to non-uniform excavation progress and site-specific geotechnical conditions.

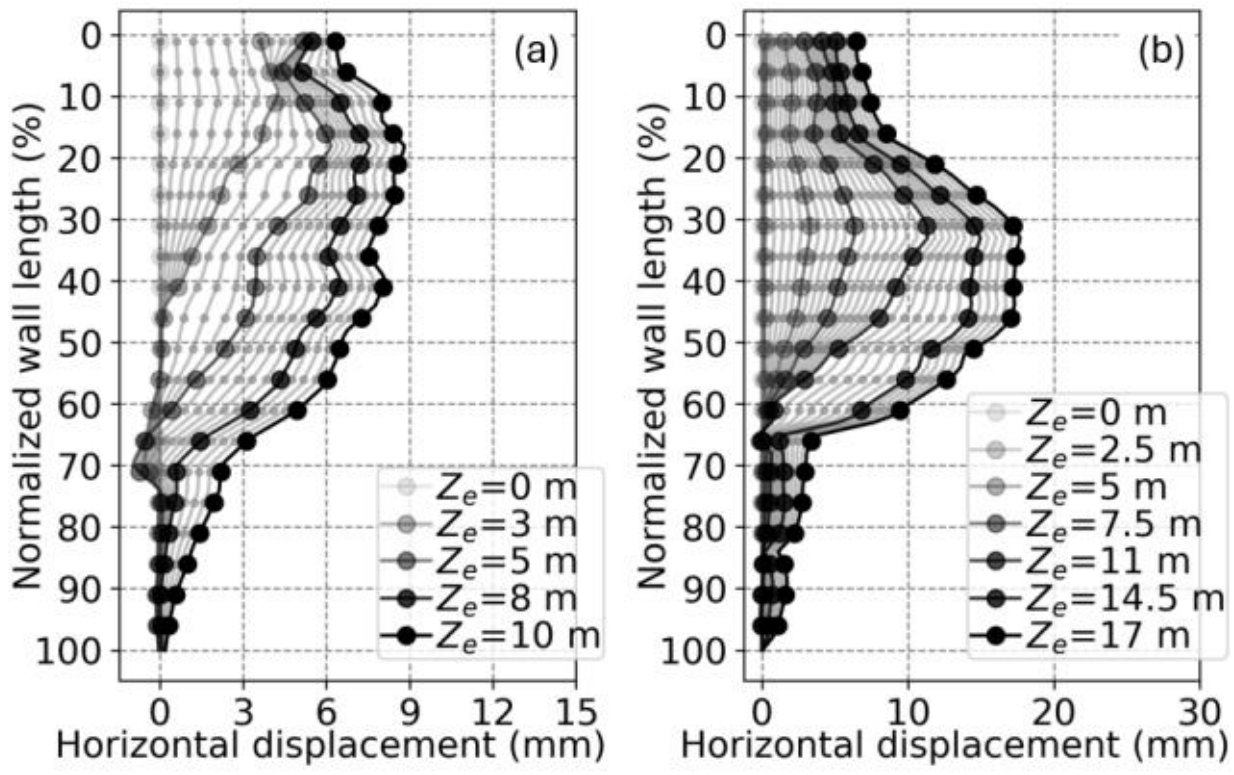

Figure 11. Field measurements for model validation at (a) Site A and (b) Site B

### 3.2.1 Overall prediction accuracy

Fig. 12 displays the ensemble model's predictions and measured displacements (i.e., actual values) considering all available target depths at each validation site. As shown in the figure, the ensemble model exhibits strong agreement with the measured data, with relatively high $R^2$ values of 0.85 for Site A and 0.90 for Site B. Most predictions are closely distributed around the 1:1 line, indicating robust and reliable predictive performance under varying excavation conditions. Although a slight tendency toward overestimation is observed at larger displacement values, the overall results demonstrate the strong practical applicability of the proposed framework.

To compare the predictive performance of the standalone models and the proposed framework, the evaluation metrics are summarized in Table 5. Overall, the proposed ensemble model consistently outperforms the standalone models at both sites. At Site A, Model C shows the best performance among the standalone models; however, the ensemble model achieves a further improvement, increasing the $R^2$ value by 0.09. Similarly, at Site B, Model A exhibits the highest performance among the standalone models, while the ensemble model again provides superior accuracy, with an $R^2$ improvement of 0.16. These results confirm the enhanced predictive capability of the proposed ensemble approach under varying site conditions.

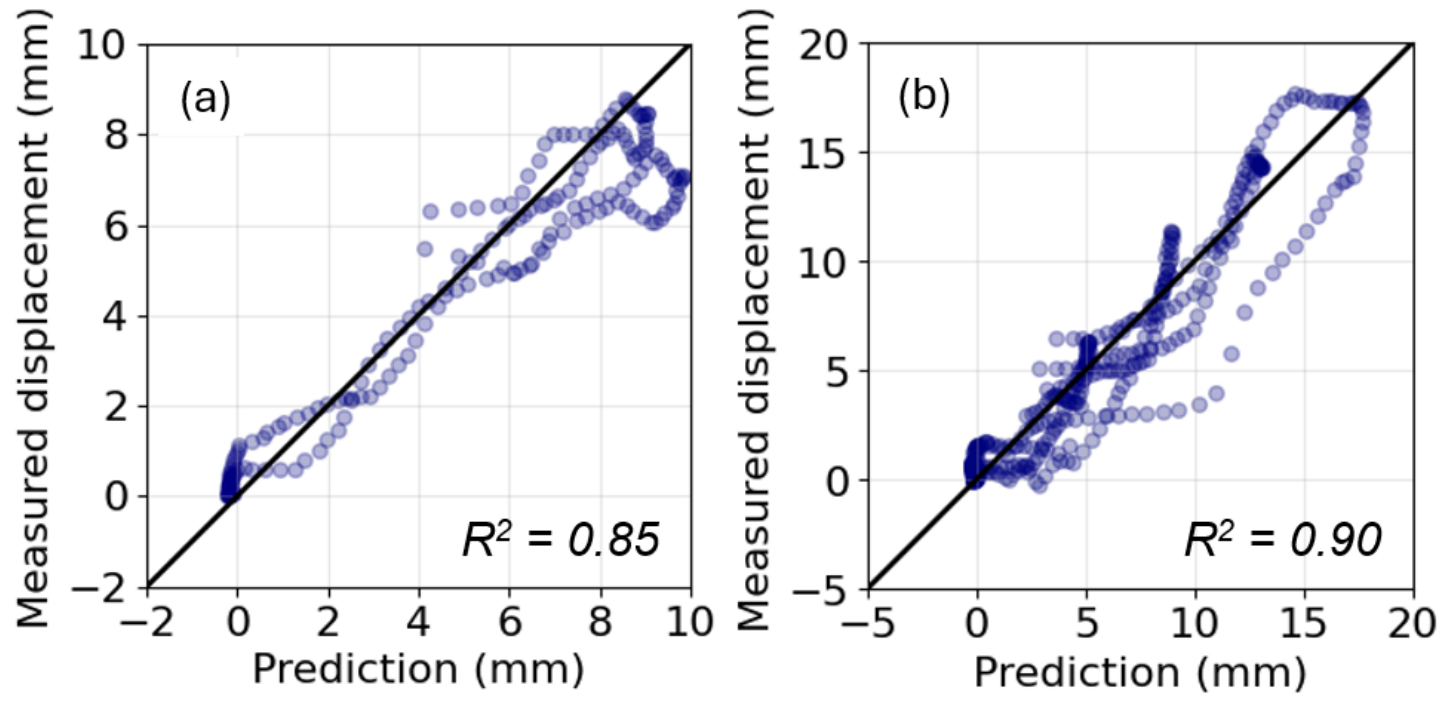

Figure 12. Comparison of ensemble model's prediction and measured displacement at (a) Site A and (b) Site B

Table 5. Site-wise prediction accuracy (MAE, RMSE, IoA, and $R^2$)

| Site | Model | MAE (mm) | RMSE (mm) | IoA | $R^2$ |
|---|---|---|---|---|---|
| A | A | 1.55 | 1.84 | 0.92 | 0.61 |
| | B | 1.83 | 2.25 | 0.88 | 0.42 |
| | C | 1.28 | 1.43 | 0.94 | 0.76 |
| | DL | 0.83 | 1.15 | 0.97 | 0.85 |
| B | A | 1.99 | 2.54 | 0.93 | 0.75 |
| | B | 2.03 | 2.53 | 0.93 | 0.75 |
| | C | 2.21 | 2.61 | 0.93 | 0.74 |
| | DL | 1.18 | 1.64 | 0.97 | 0.90 |

### 3.2.2 Site A

Fig. 13 presents detailed prediction results, illustrating how each model performs. In the left set of subplots, wall displacements measured up to an excavation depth of 5.0 m are used as input to predict the subsequent profile at 8.0 m. At this stage, the stacking ensemble (DL model) achieves an $R^2$ of 0.88, whereas the best-performing standalone model (Model C) shows a lower $R^2$ of 0.74, indicating an improvement of approximately 0.14. This demonstrates that the ensemble more accurately captures the evolving deformation shape as excavation progresses.

A similar trend is observed in the right subplots, where predictions at 10.0 m (based on measurements up to 8.0 m) yield $R^2$=0.95 for the ensemble model and $R^2$=0.87 for Model B, again confirming a consistent performance gain. Notably, while standalone models tend to either overestimate or underpredict displacement in deeper zones, the ensemble model better preserves the curvature and magnitude of the displacement profile across the entire wall depth.

However, all models consistently show limitations in accurately capturing the displacement behavior within the normalized wall length range of approximately 20–40%. This discrepancy is attributed to the fundamental difference between the training data and field measurements. Specifically, the numerical simulation data used for training exhibit relatively smooth and continuous curvature, whereas the field data reflect more irregular and non-uniform deformation patterns due to construction effects and geotechnical variability. As a result, models trained on such idealized data struggle to reproduce localized variations observed in real measurements.

Nevertheless, the ensemble model effectively compensates for these limitations by balancing the systematic prediction tendencies of the standalone models. While Models A and B tend to overestimate wall displacements and Model C tends to underestimate them, the ensemble integrates these complementary biases to produce predictions that more closely align with the actual measured behavior.

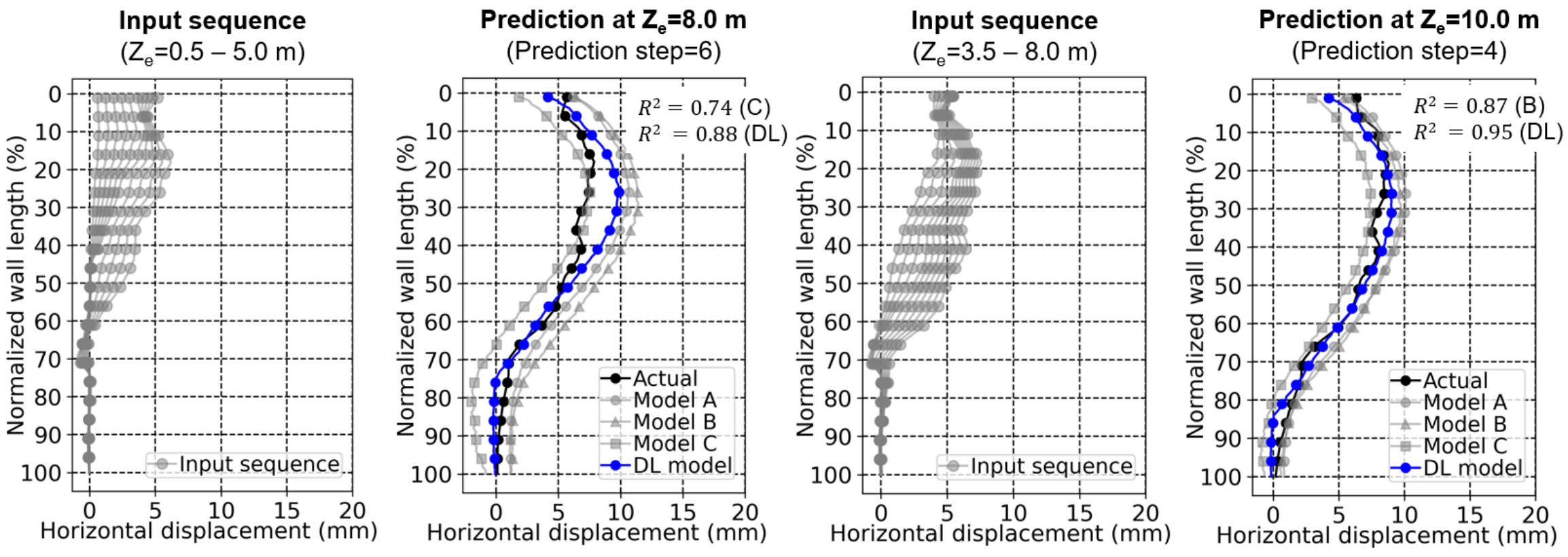

Figure 13. Detailed predictions of standalone ConvLSTM and ensemble models at Site A

### 3.2.3 Site B

Fig. 14 presents additional detailed prediction results at Site B, further demonstrating the consistency of model performance across diverse field conditions. In the first case, measurements from $Z_e$=0.5–5.0 m are used to predict the profile at $Z_e$=7.5 m, while the second case uses $Z_e$=3.0 – 7.5 m to predict $Z_e$=11.0 m. The third and fourth cases follow the same approach, using $Z_e$=6.5-11.0 m to predict $Z_e$=14.5 m and $Z_e$=10-14.5 m to predict $Z_e$=17.0 m, respectively.

Quantitatively, the ensemble model consistently achieves higher $R^2$values than the standalone models across all cases. For instance, at $Z_e$=7.5 m, the ensemble improves the $R^2$ from 0.55 (Model C) to 0.87, while at $Z_e$=11.0 m, it increases from 0.55 to 0.85. Similarly, at $Z_e$=14.5 m, the ensemble achieves $R^2$=0.93 compared to 0.82 from Model B, whereas at $Z_e$=17.0 m, a more modest improvement is observed (0.82 to 0.84). These results confirm that the ensemble model provides consistent performance gains across varying excavation conditions.

Beyond these quantitative improvements, a consistent prediction pattern is observed again across all cases: Models A and B tend to overestimate wall displacements, whereas Model C tends to underestimate them. By combining these complementary tendencies, the ensemble model effectively corrects systematic biases and produces more reliable predictions across excavation stages.

Further insight is obtained from the prediction at $Z_e$=17.0 m, where the ensemble model predicts a downward shift of the maximum displacement location as excavation progresses. This behavior reflects the model learning the redistribution of bending moments and earth pressures from the numerical simulation data. However, in actual field conditions, such behavior may not occur due to site-specific construction characteristics, including variations in support installation and construction practices. Consequently, discrepancies arise in this aspect, leading to prediction errors and highlighting the limitation of applying learned patterns from idealized data to real field conditions.

Taken together, the results from both the averaged performance and representative prediction cases for Site A and Site B indicate that the proposed ensemble model achieves high predictive accuracy, with $R^2$ values of approximately 0.85 and 0.90, respectively, in forecasting wall displacements under subsequent excavation stages. Despite the identified limitations, the ensemble framework remains capable of providing reliable predictions under practical field conditions. In the following section, a SHAP-based analysis is conducted to investigate the underlying mechanism of the ensemble framework.

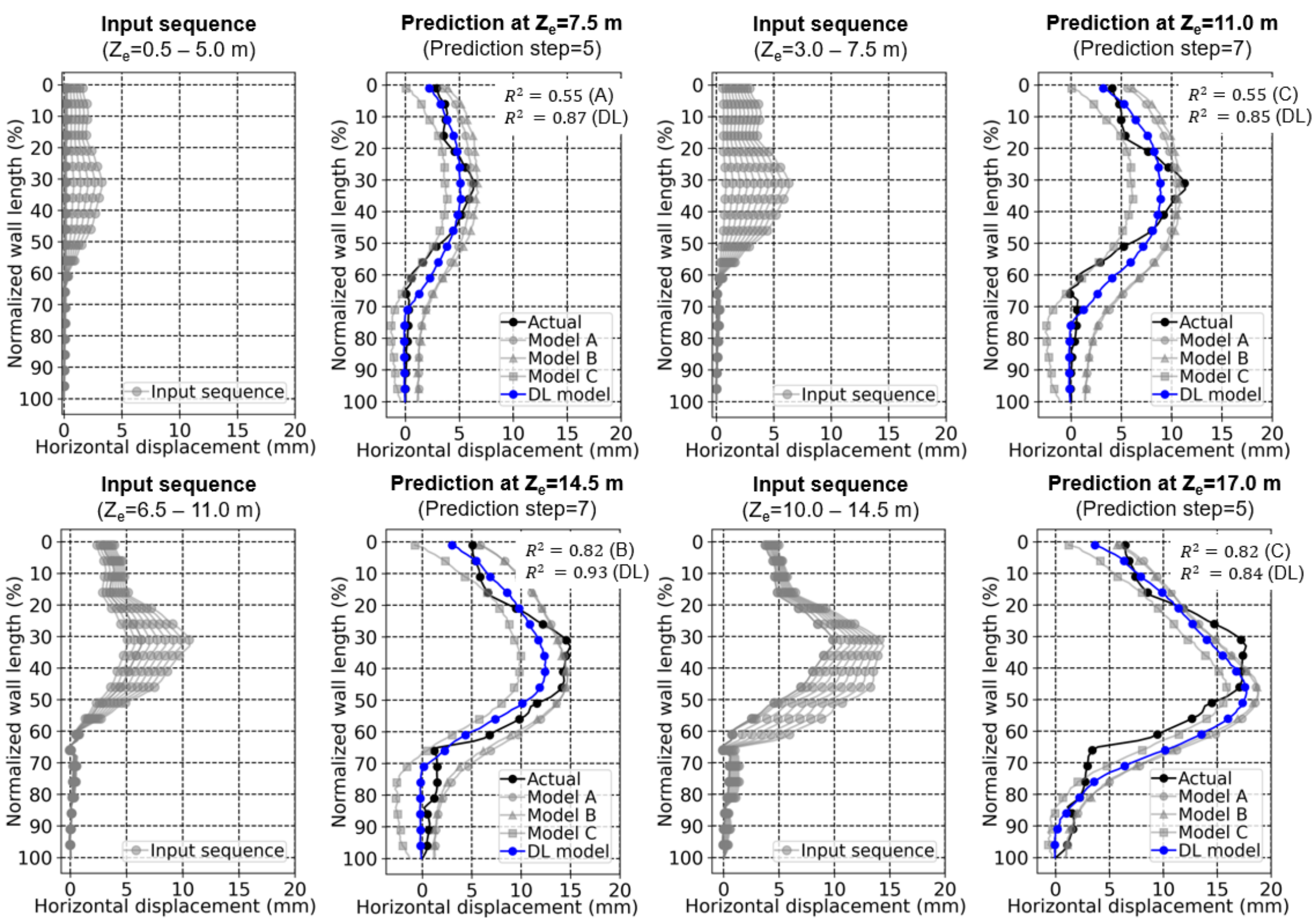

Figure 14. Detailed predictions of standalone ConvLSTM and ensemble models at Site B

## 3.3 Interpreting ensemble behavior via SHAP analysis

Table 6 summarizes the average normalized contributions of each model for the numerical and field datasets. The contributions were computed by averaging the absolute SHAP values for each model at each prediction step and normalizing them such that the sum across all models equals one. The average contribution of Model A increases from 0.327 in the numerical test data to 0.355 in the field data, while Model B shows a smaller increase from 0.255 to 0.272. In contrast, the contribution of Model C decreases from 0.416 to 0.376. These results indicate that the relative importance of shorter-resolution models (Models A and B) increases under field conditions, with Model A exhibiting a larger increase than Model B.

Fig. 15 further presents the stepwise contributions of the three standalone ConvLSTM models across all forecasting steps. Fig. 15(a) shows the results for the numerical test dataset, while Fig. 15(b) represents the average contributions obtained from the validation datasets at Sites A and B. For the numerical test data (Fig. 15(a)), the relative contributions remain relatively stable across all prediction steps, with Model C consistently exhibiting the highest contribution, followed by Models A and B. In contrast, the field validation results (Fig. 15(b)) show a different distribution of model contributions, where the relative importance of Models A and B becomes more pronounced compared to the numerical test data.

This shift can be interpreted in the context of differences between numerical simulation data and field measurements. In field conditions, excavation intervals are not uniform, and shorter intervals often result in a greater proportion of recent observations being included in the input sequence. Consequently, the influence of recent displacement changes becomes more pronounced compared to the uniformly spaced numerical data, where deformation evolves in a smoother and more consistent manner. Under such conditions, models with shorter temporal resolutions (Models A and B) tend to place greater emphasis on recent input data, which is consistent with their increased relative contributions observed in the SHAP analysis.

Overall, these results indicate that the ensemble framework effectively utilizes models with different temporal characteristics in a complementary manner. By adaptively combining models that emphasize recent changes with those that capture longer-term trends, the proposed framework achieves improved predictive performance under varying data conditions, as its nonlinear structure enables the contribution of each model to vary with the input, resulting in context-dependent amplification or suppression of individual outputs.

Table 6. Average normalized SHAP contributions of each model for numerical results and field measurements

| Model | Numerical results | Field measurements |
|---|---|---|
| A | 0.327 | 0.355 |
| B | 0.255 | 0.272 |
| C | 0.416 | 0.376 |

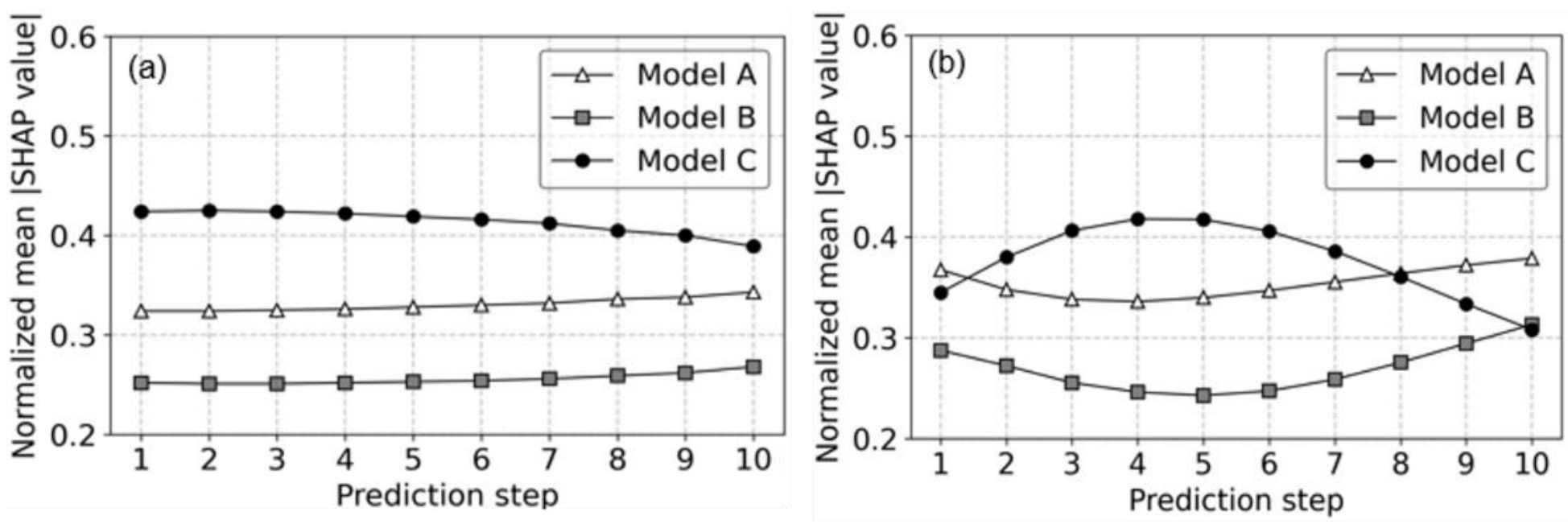

Figure 15. Stepwise relative contributions of standalone models to ensemble prediction for (a) numerical results, and (b) field measurements (both Sites A and B)

## 3.4 Model responsiveness

To further investigate the mechanisms behind the SHAP-based contribution shifts, an additional analysis was conducted using a controlled input scenario (Fig. 16). Specifically, synthetic input sequences were constructed based on measurements from Site B, where the deformation up to an excavation depth of 7.5 m was used as input, and a sudden increase in deformation at 8.0 m was introduced.

Under this setup, all models are expected to predict future deformations exceeding the most recent input. However, the models exhibit distinct behaviors: Model A responds to the recent change by predicting larger deformations, Model B produces predictions similar to the latest input, and Model C predicts smaller deformations.

This divergence can be interpreted in terms of two main factors. First, the numerical simulations used for training do not fully capture the irregular and nonlinear fluctuations present in field measurements, which may limit the models' ability to adapt to abrupt changes. Second, each model's input window length influences its responsiveness. Model A (resolution=3) tends to respond more strongly to recent inputs, whereas Model C (resolution=10) places greater emphasis on long-term trends and may smooth out sudden changes. As a result, Model C may be less sensitive to abrupt deviations, which can lead to early-step errors that propagate in recursive forecasting.

This behavior may help interpret the SHAP-based contribution patterns observed in the field validation. In real-world applications, where irregular and abrupt changes frequently occur, models with shorter input windows (e.g., Model A) tend to capture recent variations more effectively. Consequently, their predictions can exert a larger influence on the ensemble output, resulting in higher SHAP values. Conversely, models with longer input windows (e.g., Model C), which emphasize smoother long-term trends, may be less responsive to sudden changes and therefore exhibit smaller contributions under such conditions. These observations are consistent with the SHAP results, where short-resolution models tend to show relatively higher contributions in field scenarios.

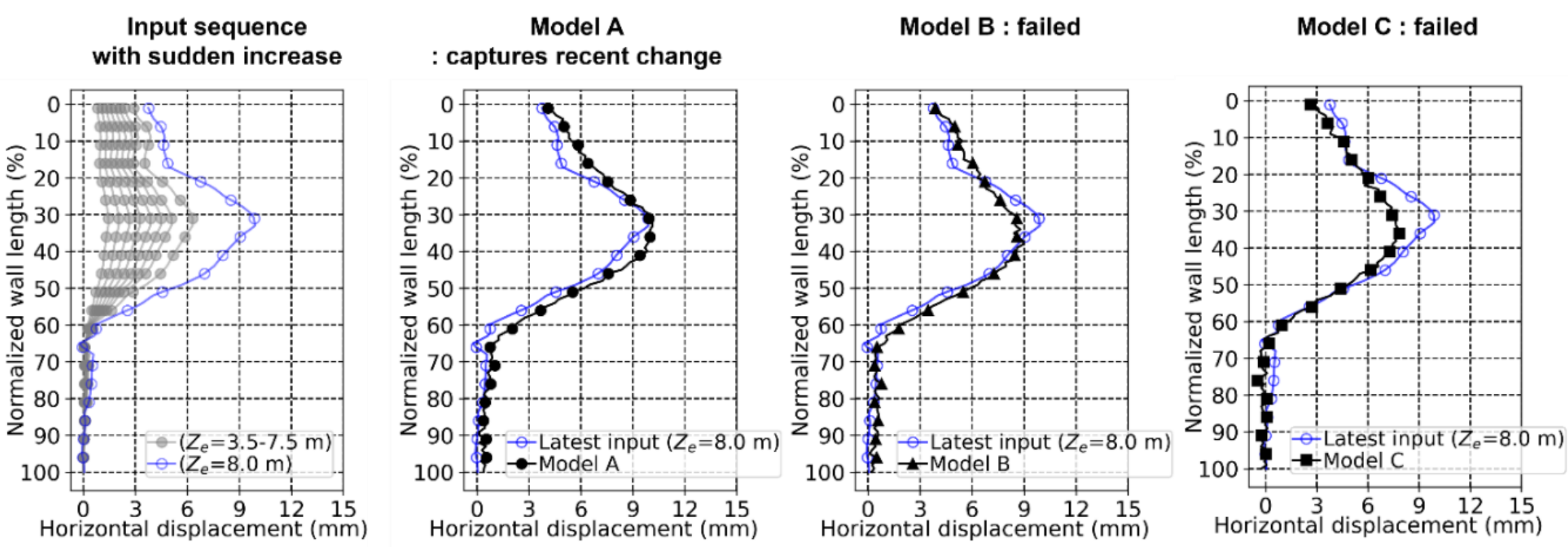

Figure 16. Comparison of model responses to a sudden deformation increase in the latest input

## 3.4 Bias-variance decomposition via Monte-Carlo dropout

Fig. 17 illustrates the decomposition results of error bias and variance for the standalone models and the ensemble model, estimated using Monte Carlo (MC) dropout. Figs. 17(a) and (b) show the error bias and variance for numerical simulation data, respectively, while Figs. 17(c) and (d) present the corresponding results for field measurements (Sites A and B). For each prediction step, 100 stochastic forward passes were performed, and the prediction error was defined as the difference between predictions and ground truth (i.e., measured displacement). The error bias and variance were then computed as the mean and the variance of the errors across MC samples, respectively.

Comparing Figs. 17(a) and (c), the standalone models exhibit consistent and systematic bias patterns. Models A and B show overestimation, whereas Model C shows underestimation, consistent with previously observed trends. This suggests that the bias patterns remain consistent when the framework is applied to field data. In contrast, the ensemble model maintains bias values close to zero in both cases, confirming that it effectively compensates for opposing systematic errors.

Regarding variance, all standalone models exhibit increasing variance with prediction step in both numerical and field cases, which is consistent with the effect of error accumulation in recursive forecasting. However, notable differences are observed between the two datasets. In the numerical results (Fig. 17(b)), the variance tends to stabilize or slightly decrease as the prediction horizon extends. In contrast, in the field measurements (Fig. 17(d)), the variance remains relatively stable in early steps and then gradually increases.

In both numerical and field measurement results, the ensemble model exhibits relatively higher variance at the initial prediction steps, which is likely related to the meta-model structure under MC dropout. Specifically, the relatively high dropout rate introduces stochastic masking of neurons during inference, so the nonlinear combination of base-model outputs is repeatedly perturbed. When base predictions still differ at early steps, this masking can amplify differences across stochastic passes, yielding a wider spread of predicted errors.

Taken together, the results indicate that the ensemble model effectively reduces systematic bias while stabilizing variance growth, demonstrating consistent advantages across both numerical simulations and field measurements.

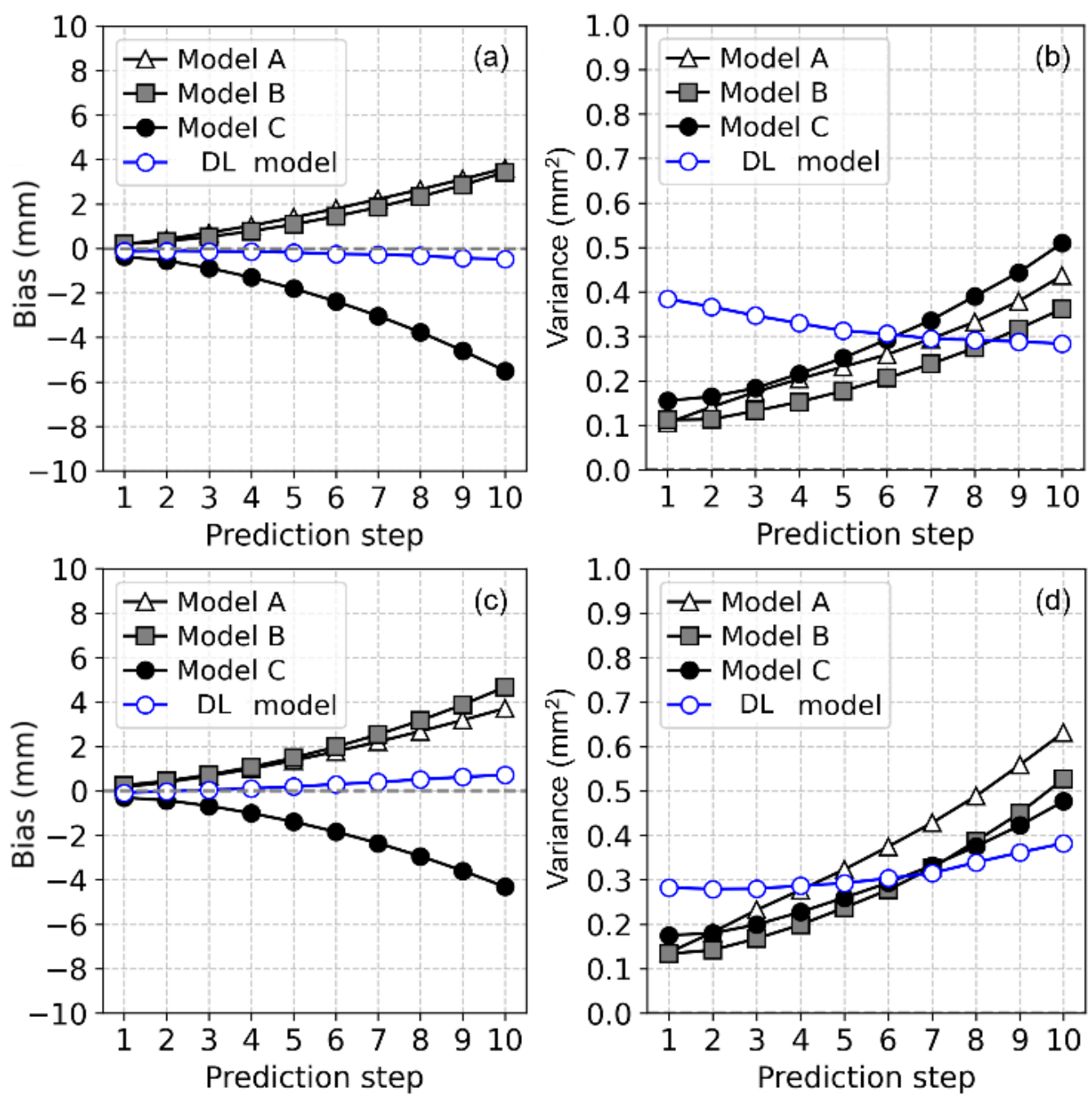

Figure 17. Bias-variance decomposition results: (a) bias and (b) variance for numerical results; (c) bias and (d) variance for field measurements

## 4. Conclusions

In this study, we developed a multi-resolution ConvLSTM ensemble framework to improve long-horizon forecasting of retaining structure behavior during staged excavation. We generated a comprehensive numerical dataset via PLAXIS2D simulations, sampling geotechnical and structural parameters stochastically to capture site variability. Multiple ConvLSTM models, each configured with a different temporal resolution, were integrated using a stacking ensemble approach. The resulting framework was evaluated on both synthetic simulation results and real-world field measurements. The key findings are as follows:

- The proposed framework substantially improves predictive accuracy while effectively mitigating error accumulation, resulting in a longer and more reliable prediction horizon for retaining wall displacements across both numerical results and field measurements.
- In the evaluation of model applicability using field measurements, the ensemble model achieved an average $R^2$ of 0.85 and 0.90 at Sites A and B, respectively, whereas the standalone ConvLSTM models showed pronounced error accumulation, with average $R^2$ values dropping below 0.75 at the same sites.
- The stacking ensemble model benefited from combining ConvLSTM models with different temporal resolutions, and SHAP-based analysis of relative contributions indicates that, under field conditions, higher-resolution models tend to contribute more, while lower-resolution models contribute less.

- Based on the MC dropout–based analysis of error bias and variance, the proposed framework tends to reduce systematic bias while moderating and stabilizing the growth of variance over the prediction horizon.
- Despite being trained on an idealized FEM-derived dataset, the proposed framework effectively captures nonlinear deformation patterns in real-world conditions, suggesting its potential for practical field applicability.

While the results demonstrate meaningful improvements in predictive performance, several limitations remain. The proposed framework has been validated under limited numerical conditions and a small number of field cases; therefore, further validation across diverse excavation scenarios and geotechnical conditions is required. In addition, application to different field conditions may require adjustments to the prediction approach or optimization of the model structure. Furthermore, the numerical dataset does not fully capture time-dependent soil behaviors (e.g., consolidation and creep) or the irregular and highly variable deformation patterns observed in real-world conditions. Accordingly, efforts to reduce the domain gap and improve adaptability to field data are necessary. Future work should focus on broader field validation and enhancing the applicability and robustness of the framework through the integration of in situ monitoring data and appropriate model adaptation strategies.

## Declaration of Competing Interest

The authors declare that they have no known competing financial interests or personal relationships that could have appeared to influence the work reported in this paper.

## Acknowledgements

This work was supported by the National Research Foundation of Korea (NRF) grant funded by the Korea government (MSIT) (No. 2023R1A2C1007635).